\documentclass[10pt,twocolumn,letterpaper]{article}

\usepackage{wacv}
\usepackage{times}
\usepackage{epsfig}
\usepackage{graphicx}
\usepackage{amsmath}
\usepackage{amssymb}

\usepackage[utf8]{inputenc}
\usepackage[percent]{overpic}
\usepackage[table]{xcolor}
\usepackage{multirow}
\usepackage{color}
\usepackage{enumitem}

\makeatletter
\renewcommand{\paragraph}{%
  \@startsection{paragraph}{4}%
  {\z@}{1ex \@plus 1ex \@minus .2ex}{-1em}%
  {\normalfont\normalsize\bfseries}%
}
\makeatother

\makeatletter
\let\MYcaption\@makecaption
\makeatother

\usepackage[font=footnotesize]{subcaption}

\makeatletter
\let\@makecaption\MYcaption
\makeatother

\usepackage[pagebackref=true,breaklinks=true,colorlinks,bookmarks=false]{hyperref}

\urldef{\supplementary}\url{https://youtu.be/7fsXq5EA0Rw}

\usepackage[sort&compress,capitalize,noabbrev]{cleveref}
\creflabelformat{equation}{#2#1#3}

\usepackage{glossaries}

\newacronym{wht}{WHT}{Walsh-Hadamard-Transform}
\newacronym{sgm}{SGM}{Semi-Global-Matching}
\newacronym{prsm}{PRSM}{Piece-wise Rigid Scene Model}
\newacronym{sed}{SED}{Structured Edge Detection}
\newacronym{sor}{SOR}{Successive Over-Relaxation}

\newcommand\Tstrut{\rule{0pt}{2.6ex}}         
\newcommand\Bstrut{\rule[-0.8ex]{0pt}{0pt}}   

\wacvfinalcopy 

\def\wacvPaperID{18} 

\ifwacvfinal\fi
\begin{document}

\title{SceneFlowFields: Dense Interpolation of Sparse Scene Flow Correspondences}

\author{René Schuster\textsuperscript{1} \hspace{0.5cm} Oliver Wasenmüller\textsuperscript{1} \hspace{0.5cm} Georg Kuschk\textsuperscript{2} \hspace{0.5cm} Christian Bailer\textsuperscript{1} \hspace{0.5cm} Didier Stricker\textsuperscript{1} \\
\textsuperscript{1}DFKI - German Research Center for Artificial Intelligence \hspace{5mm}
\textsuperscript{2}BMW Group \\
{\tt\small firstname.lastname@\string{dfki,bmw\string}.de}
}

\maketitle
\ifwacvfinal\thispagestyle{empty}\fi

\begin{abstract}
   While most scene flow methods use either variational optimization or a strong rigid motion assumption, we show for the first time that scene flow can also be estimated by dense interpolation of sparse matches. To this end, we find sparse matches across two stereo image pairs that are detected without any prior regularization and perform dense interpolation preserving geometric and motion boundaries by using edge information. A few iterations of variational energy minimization are performed to refine our results, which are thoroughly evaluated on the KITTI benchmark and additionally compared to state-of-the-art on MPI Sintel. For application in an automotive context, we further show that an optional ego-motion model helps to boost performance and blends smoothly into our approach to produce a segmentation of the scene into static and dynamic parts.
\end{abstract}

\section{Introduction} \label{sec:intro}
Scene flow describes the perceived 3D motion field with respect to the observer. It can thereby be considered as an extension to optical flow, which in comparison is the apparent motion field in 2D image space. Many applications such as robot navigation, high level vision tasks, \eg moving object detection \cite{rabe2007fast}, and driver assistance systems rely on an accurate motion estimation of their surroundings. Especially the latter ones have great potential to make traffic more comfortable and much safer. While scene flow is not only a more detailed representation of the real world motion compared to optical flow, scene flow algorithms also reconstruct the 3D geometry of the environment. Due to the increased complexity when compared to depth or optical flow estimation, scene flow has only recently become of bigger interest. At the same time, simply estimating depth and optical flow separately to obtain scene flow (see \cite{lenz2011sparse,schuster2018combining,schuster2017towards,taniai2017fsf}) is not exploiting the full potential of the underlying data. The splitting produces incoherent results, limits the exploitation of inherent redundancies, and in general yields a non-dense scene flow field. Approaches that combine stereo depth estimation, \eg \cite{hirschmuller2008SGM}, and 2D optical flow, \eg \cite{brox2011large,sun2014quantitative}, have clearly been outperformed by a huge margin in the KITTI benchmark \cite{geiger2012kitti}.

\begin{figure}[t]
	\begin{center}
		\begin{subfigure}[c]{0.95\columnwidth}
			\includegraphics[width=1\textwidth]{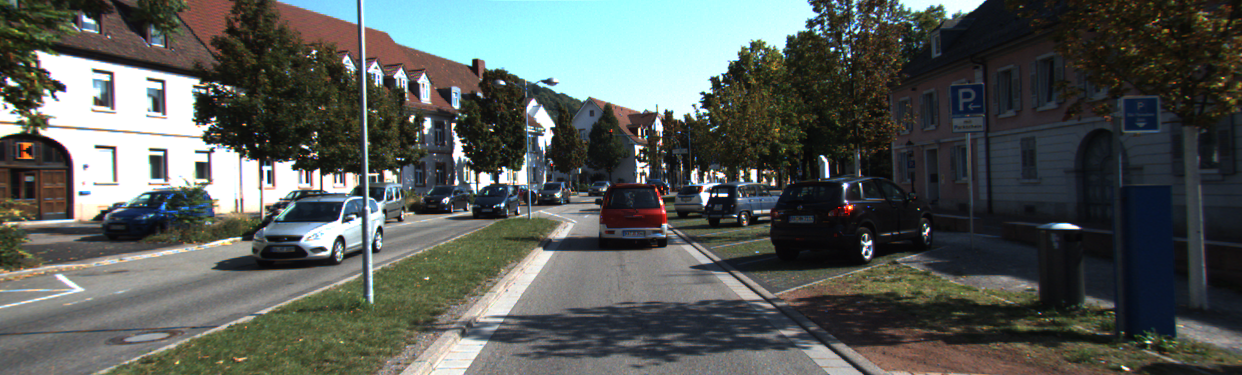}%
			\raisebox{2pt}{\makebox[0pt][r]{\textcolor{white}{\bf a)~}}}%
			\vspace{1mm}%
		\end{subfigure}
		\begin{subfigure}[c]{0.95\columnwidth}
			\includegraphics[width=1\textwidth]{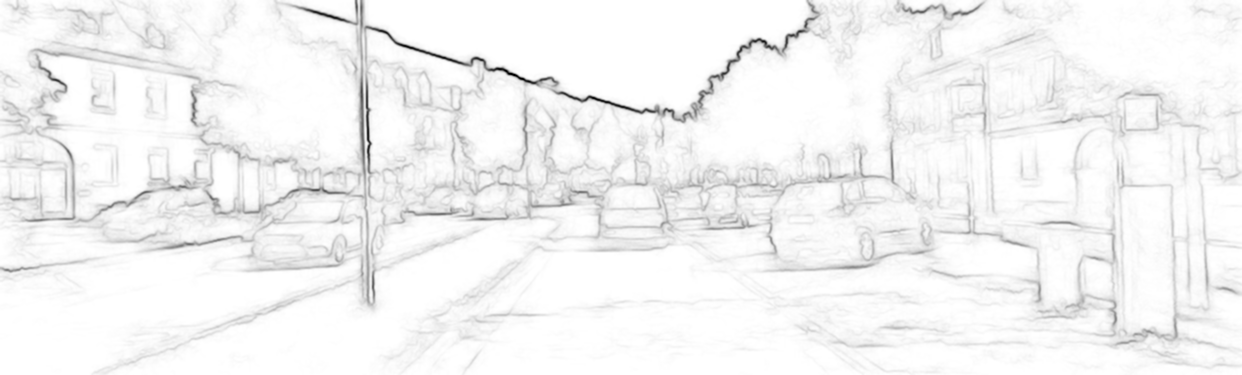}%
			\raisebox{2pt}{\makebox[0pt][r]{\bf b)~}}%
			\vspace{1mm}%
		\end{subfigure}
		\begin{subfigure}[c]{0.95\columnwidth}
			\includegraphics[width=1\textwidth]{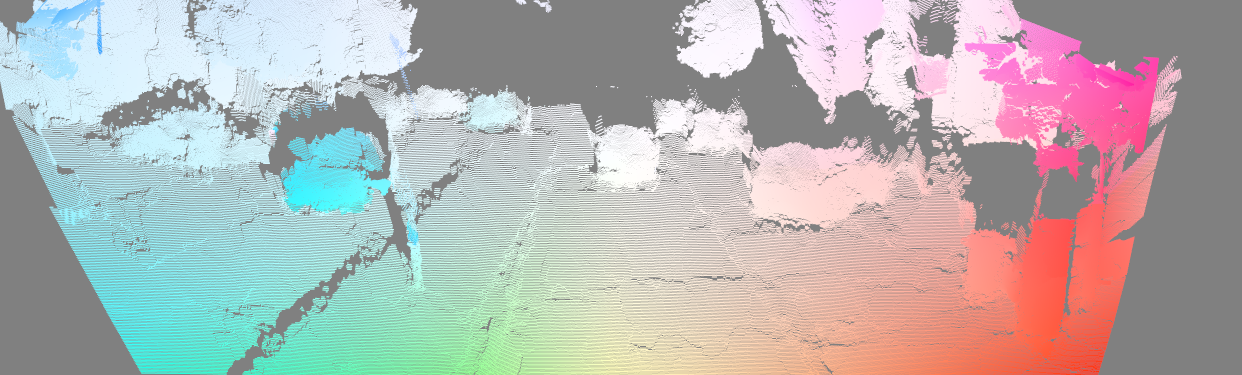}%
			\raisebox{2pt}{\makebox[0pt][r]{\bf c)~}}%
			\vspace{1mm}%
		\end{subfigure}
		\begin{subfigure}[c]{0.95\columnwidth}
			\includegraphics[width=1\textwidth]{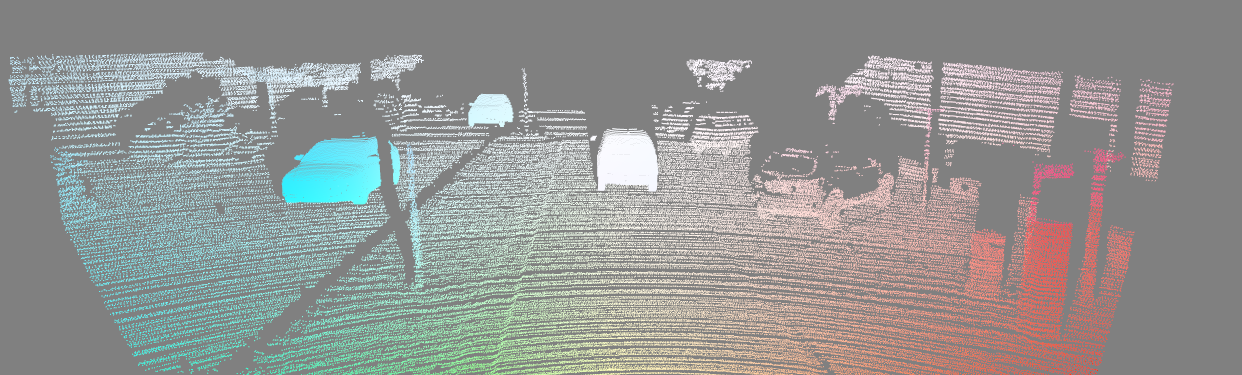}%
			\raisebox{2pt}{\makebox[0pt][r]{\bf d)~}}%
		\end{subfigure}
	\end{center}
	\caption{We present SceneFlowFields which uses stereo image pairs {\bf (a)}, extracts scene flow boundaries {\bf (b)}, and computes a dense scene flow field {\bf (c)} that we compare to ground truth {\bf (d)}. The color of the point clouds encodes the optical flow.}
	\label{fig:teaser}
\end{figure}

\begin{figure}[t]
\begin{center}
	\includegraphics[width=0.95\columnwidth]{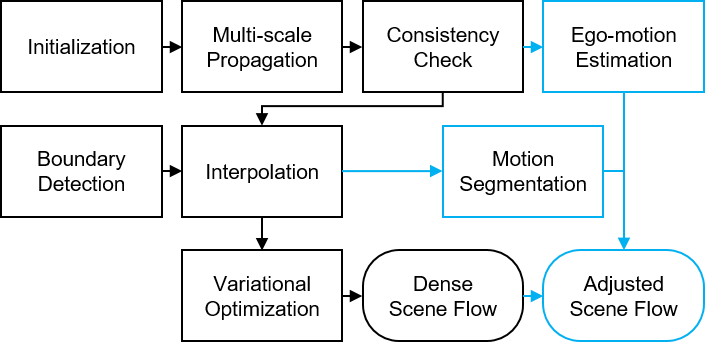}
\end{center}
\caption{Overview of our SceneFlowFields. Blue color indicates the optional ego-motion extension.}
\label{fig:overview}
\end{figure}

Most state-of-the-art approaches are either designed for indoor scenarios or describe outdoor scenes under mostly stationary or rigid motion assumptions. Contrary to that, our method is very versatile. In fact, we do not employ any a-priori regularization. Our matching process inherently encodes a first order local smoothness assumption and is itself solely data term based, while our interpolation scheme allows for very sharp discontinuities in the scene flow field. An optional ego-motion extension that incorporates additional assumptions improves accuracy for challenging traffic data sets like KITTI, but is no mandatory part of our method. The basic version of the method and its optional extension are illustrated in \cref{fig:overview}. The single stages of our approach are visualized in the supplementary video\footnote{\supplementary}.

In more detail, we present a new scene flow approach -- called SceneFlowFields -- that densely interpolates sparse scene flow matches. The interpolation preserves boundaries of the 3D geometry and of moving objects by edge-preserving interpolation based on an improved edge detector to approximate scene flow boundaries. Matches are obtained by multi-scale propagation with random search to compute a dense scene flow field that in turn is filtered to remove outliers and leave sparse, robust correspondences across all images. The combination of multi-scale matching and edge-preserving interpolation sums to a novel method which is in strong contrast with any existing method that estimates scene flow. Matching can optionally be used to estimate ego-motion and obtain sparse motion indicators which can be interpolated to a dense motion segmentation. Optionally using ego-motion, we can reconstruct the scene flow for static parts of the scene directly. However, our method does not rely on ego-motion estimation. In particular, our contribution consists of:
\begin{itemize}[noitemsep,topsep=1pt,label={\tiny\raisebox{0.75ex}{\textbullet}}]
\item A novel method to find scene flow matches.
\item A new interpolation method for scene flow that preserves boundaries of geometry and motion.
\item An improved edge detector to approximate scene flow boundaries.
\item An optional approach for straightforward integration of ego-motion.
\item Thorough evaluation of our method on KITTI and MPI Sintel with comparison to state-of-the-art methods.
\end{itemize}

\section{Related Work} \label{sec:relatedwork}
Starting from Vedula \etal \cite{vedula1999three} who was among the first to compute 3D scene flow, many variational approaches followed. First using pure color images as input \cite{basha2013multi,huguet2007variational} and later using RGB-D images \cite{herbst2013rgbd,jaimez2015primal,wedel2008efficient}. While a variational formulation is typically complex, \cite{jaimez2015primal} achieved real-time performance with a primal-dual framework. Yet, all these approaches are sensitive to initialization and can not cope with large displacements \cite{wasenmueller2017towards}, which is why they use a coarse-to-fine scheme. That in turn tends to miss finer details. Furthermore, the RGB-D approaches rely on depth sensors that either perform poorly in outdoor scenarios or are accordingly very expensive \cite{yoshida2017time}.

Since it is hard to capture ground truth scene flow information, there exist only very few data sets to evaluate scene flow algorithms on. Most of them use virtually rendered scenes to obtain the ground truth data \cite{butler2012sintel,gaidon2016vkitti,mayer2016large}. To the best of our knowledge the only realistic data set that provides a benchmark for scene flow is the KITTI Vision Benchmark \cite{geiger2012kitti} that combines various tasks for automotive vision. Its introduction has played an important role in the development of stereo and optical flow algorithms, and the extension by \cite{menze2015object} has also driven the progress in scene flow estimation.

Due to the advent of a piece-wise rigid plane model \cite{vogel2013PRSF}, scene flow has recently achieved a boost in performance. The majority of top performing methods at the KITTI Vision Benchmark employ this model to enforce strong regularization \cite{lv2016CSF,menze2015object,vogel2015PRSM}. 
In \cite{vogel2015PRSM} the authors encode this model by alternating assignment of each pixel to a plane segment and each segment to a rigid motion, based on a discrete set of planes and motions. In \cite{menze2015object} the complexity of the model is further lowered by the assumption that a scene consists of only very few independently moving rigid objects. Thus each plane segment only needs to be assigned to one object. All segments assigned to the same object share the same motion. By propagation of objects over multiple frames, \cite{neoral2017object} achieves temporal consistency for \cite{menze2015object}. The authors of \cite{lv2016CSF} solve the pixel-to-plane assignment and the plane-to-motion assignment in a continuous domain.

Another promising strategy builds on the decomposition of a scene into static and moving parts \cite{taniai2017fsf}. While the motion of dynamic objects is estimated by solving a discrete labeling problem (as in \cite{chen2016full}) using the \gls*{sgm} \cite{hirschmuller2008SGM} algorithm, the perceived motion of all static parts is directly obtained from the 3D geometry of the scene and the ego-motion of the camera. This approach is especially convenient for scenes, where only a small proportion consists of moving objects, like it is typically the case in traffic scenarios.
However, any a-priori assumption limits the versatility of a method. A rigid plane model performs poorly when applied to deformable objects, and ego-motion estimation for highly dynamic scenes is hard.

Our scene flow method differs from any of the mentioned approaches. We find sparse scene flow matches that are interpolated to a dense scene flow field, recovering the geometry of the scene and the 3D motion. 
Our method has to be distinguished from purely variational approaches. Although we use variational optimization, it can be considered as a post-processing step for refinement.
During interpolation, we assume that the geometry of a scene can be modeled by small planar segments, but we do not initially presume any segmentation. In fact, the size of our plane segments only depends on the density of the matches, which leads to smoothly curved shapes where matches are dense and to planar patches where matches are sparse. The same holds for our piece-wise affine motion model that is used to interpolate the 3D motion. These differences in the model and additionally the difference in the optimization method draw a clear boundary between our method and \cite{lv2016CSF,menze2015object,vogel2015PRSM}. 
If we apply our optional ego-motion model there is a conceptual overlap to \cite{taniai2017fsf} which also uses ego-motion to estimate the 3D motion for the static parts of the scene. However, if we do not apply the ego-motion model, both methods have no noteworthy similarities. In any case, the way we estimate the ego-motion and the motion segmentation differs essentially from \cite{taniai2017fsf}.

Finally, one has to differentiate between dual-frame \cite{lv2016CSF,menze2015object,vogel2015PRSM} and multi-frame \cite{neoral2017object,taniai2017fsf,vogel2015PRSM} approaches. Especially the images of KITTI have several characteristics that make matching between two frame pairs much more challenging than in a multi-view setting. First, considerably large stereo and flow displacements. Second, difficult lighting conditions and many reflective surfaces. Third, fast ego-motions combined with a low frame rate, which causes large regions to move out of image bounds. This has to be kept in mind when comparing results across these two categories.

\section{SceneFlowFields} \label{sec:method}
For scene flow computation we assume to have the typical stereo image information provided, \ie two rectified stereo image pairs ($I_l^0, I_l^1, I_r^0, I_r^1$) at times $t_0$ and $t_1$ along with the camera intrinsics. We further assume that the baseline $B$ is known. For rectified images, the baseline describes the relative pose between the left and right cameras as translation parallel to the image plane. 
We represent scene flow as a 4D vector $\textbf{u} =(u,v,d_0,d_1)^T$ consisting of two optical flow components $u$, $v$ and the disparity values $d_0$, $d_1$ for both time steps. During matching, we jointly optimize all four components to obtain coherent scene flow.
Given the mentioned information, we estimate a dense scene flow field as follows: For $k$ subscales we initialize the coarsest by finding the best correspondences from kD-trees build with feature vectors using \gls*{wht} \cite{hel2005real}. For all $k+1$ scales (the $k$ subscales plus full resolution), we iteratively propagate scene flow vectors and adjust them by random search. Afterwards, the dense scene flow map on full resolution is filtered using an inverse scene flow field and a region filter. The filtered scene flow map is further thinned out by only taking the best match in each non-overlapping $3 \times 3$ block. Scene flow boundaries are detected using a structured random forest. Geometry and 3D motion are separately interpolated based on a boundary-aware neighborhood. Finally, we refine the 3D motion by variational optimization. An overview is outlined in \cref{fig:overview}. 

\subsection{Sparse Correspondences} \label{sec:matching}
\paragraph*{Matching Cost.}
The matching cost in our algorithm solely depends on a data term. No additional smoothness assumptions are made like \eg in \cite{herbst2013rgbd,huguet2007variational,lv2016CSF,menze2015object,vogel2013PRSF,vogel2015PRSM}. Given a scene flow vector, we define its matching cost by the sum of Euclidean distances between SIFTFlow features \cite{liu2011siftflow} over small patches for three image correspondences. These correspondences are the stereo image pair at time $t_0$, the temporal image pair for the left view point (standard optical flow correspondence) and a cross correspondence between the reference frame and the right frame at the next time step. This leads to the following cost $C$ for a scene flow vector $\textbf{u}$ at pixel $p$:
\begin{equation} \label{eq:matchingcost}
\begin{split}
	C &= \sum_{\tilde{p} \in W(p)} \left\lVert \phi\left(I_l^0\left(\tilde{p}\right)\right) - \phi\left(I_l^1\left(\tilde{p}+(u,v)^T\right)\right) \right\rVert + \\
	&\quad \left\lVert \phi\left(I_l^0\left(\tilde{p}\right)\right) - \phi\left(I_r^0\left(\tilde{p}+(-d_0,0)^T\right)\right) \right\rVert + \\
	&\quad \left\lVert \phi\left(I_l^0\left(\tilde{p}\right)\right) - \phi\left(I_r^1\left(\tilde{p}+(u-d_1,v)^T\right)\right) \right\rVert
\end{split}
\end{equation}
$W(p)$ is a $7 \times 7$ patch window centered at pixel $p$ and $\phi\left(I\left(p\right)\right)$ returning the first three principal components of a SIFT feature vector for image $I$ and pixel $p$. The principal axes are computed for the combined SIFT features of all four images. At image boundaries we replicate the boundary pixel to pad images.

\paragraph*{Initialization.} 
Initialization is based on kD-trees similar to \cite{he2012computing}, but with three trees, using \gls*{wht} features as in \cite{bailer2015flow,wannenwetsch2017probflow}. For each frame other than the reference frame, we compute a feature vector per pixel and store them in a tree. To initialize a pixel of the reference image, we query the feature vector of the pixel to the pre-computed kD-trees. Scene flow matches are then obtained by comparing all combinations of the leafs for each queried node according to the matching data term introduced before (\cref{eq:matchingcost}). Since our stereo image pairs are rectified, for the images observed from the right camera view, we create kD-trees which regard the epipolar constraint, \ie queries for such a tree will only return elements, which lie on the same image row as the query pixel. This way, we can efficiently lower the number of leaves per node for the epipolar trees, which speeds up the initialization process without loss of accuracy. For further acceleration, we use this initialization on the coarsest resolution only, and let the propagation fill the gaps when evolving to the next higher scale.

\paragraph{Multi-Scale Propagation.} 
The initial matches get spread by propagation and steadily refined by random search. This is done over multiple scales which helps to distribute rare correct initial matches over the whole image. For each scale, we run several iterations of propagation in one out of the four image quadrants so that each direction is used equally. During propagation, a scene flow vector is replaced if the propagated vector has a smaller matching cost. If this is not the case, the propagation along this path continues with the existing scene flow vector. After each iteration we perform a random search. That means that for all pixels we add a uniformly distributed random offset in the interval $\left]-1,1\right[$ in pixel units of the current scale to each of the four scene flow components and check whether the matching cost decreases. Both propagation and random search help to obtain a smoothly varying vector field and to find correct matches even if the initialization is slightly flawed. For the different scale spaces, we simulate the scaling by smoothing the images and taking only every $n$-th pixel for a subsampling factor of $n = 2^k$ so that the patches consist of the same number of pixels for all scales. This way, we prevent (up)sampling errors because all operations are performed on exact pixel locations on the full image resolution. Smoothing is done by area-based downsampling followed by upsampling using Lanczos interpolation. Note that this matching method has already been used in \cite{bailer2015flow}, but while they use it for optical flow, we apply it to twice as many dimensions in search space.

\begin{figure}[t]
	\begin{center}
		\begin{subfigure}[c]{0.475\columnwidth}
			\includegraphics[width=1\textwidth]{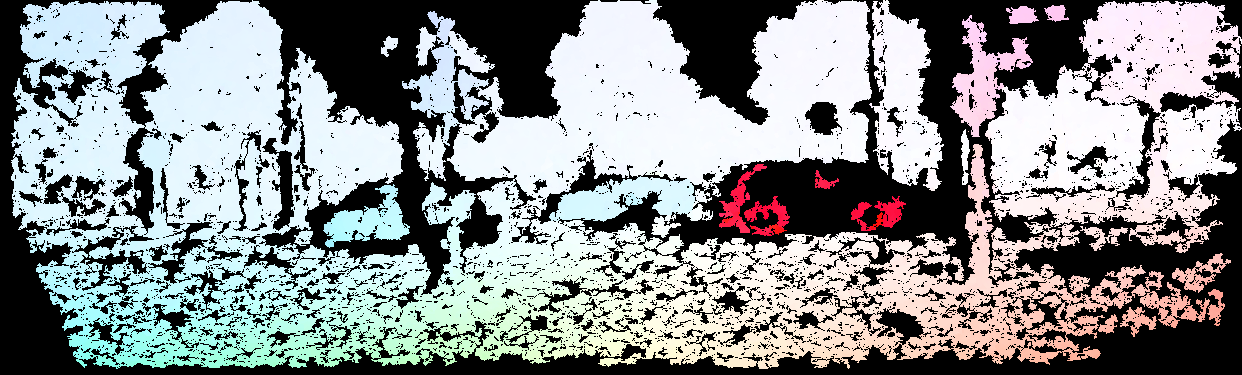}%
			\vspace{0.5mm}%
		\end{subfigure}
		\begin{subfigure}[c]{0.475\columnwidth}
			\includegraphics[width=1\textwidth]{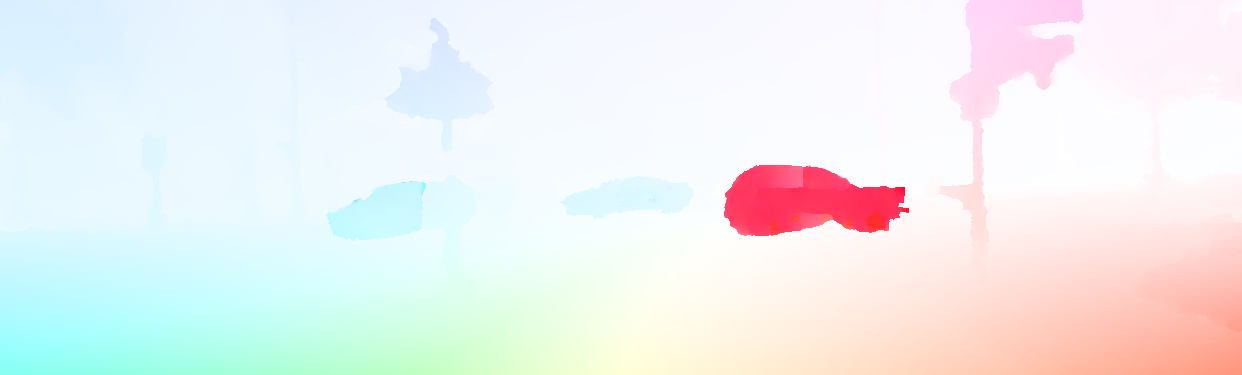}%
			\raisebox{2pt}{\makebox[0pt][r]{\bf a)~}}%
			\vspace{0.5mm}%
		\end{subfigure}
		\begin{subfigure}[c]{0.475\columnwidth}
			\includegraphics[width=1\textwidth]{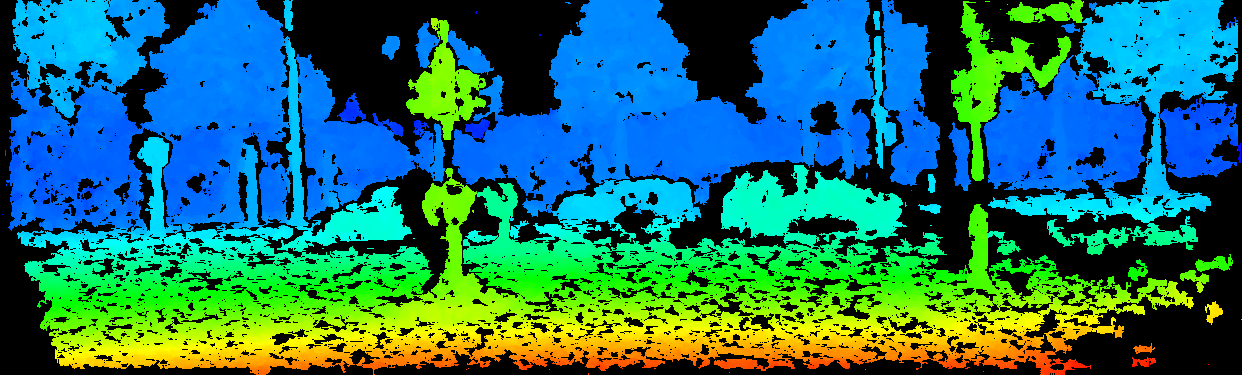}%
			\vspace{0.5mm}%
		\end{subfigure}
		\begin{subfigure}[c]{0.475\columnwidth}
			\includegraphics[width=1\textwidth]{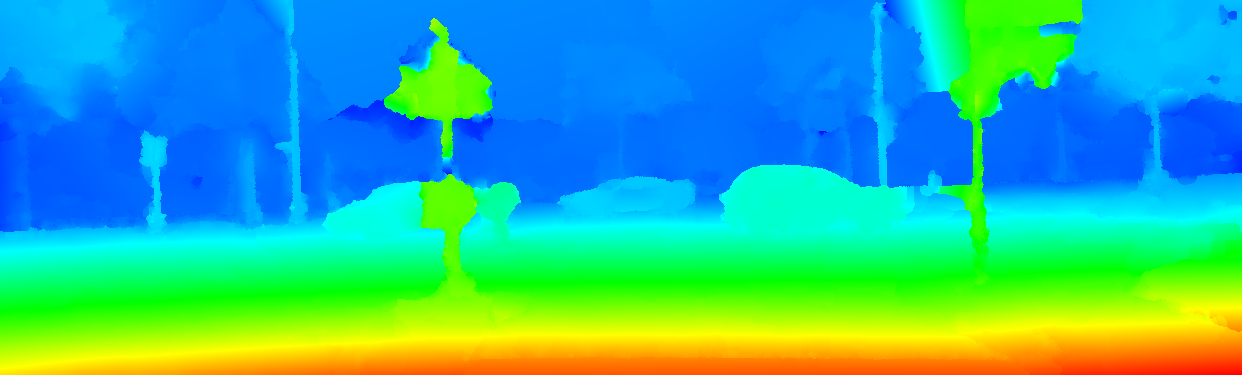}%
			\raisebox{2pt}{\makebox[0pt][r]{\bf b)~}}%
			\vspace{0.5mm}%
		\end{subfigure}
		\begin{subfigure}[c]{0.475\columnwidth}
			\includegraphics[width=1\textwidth]{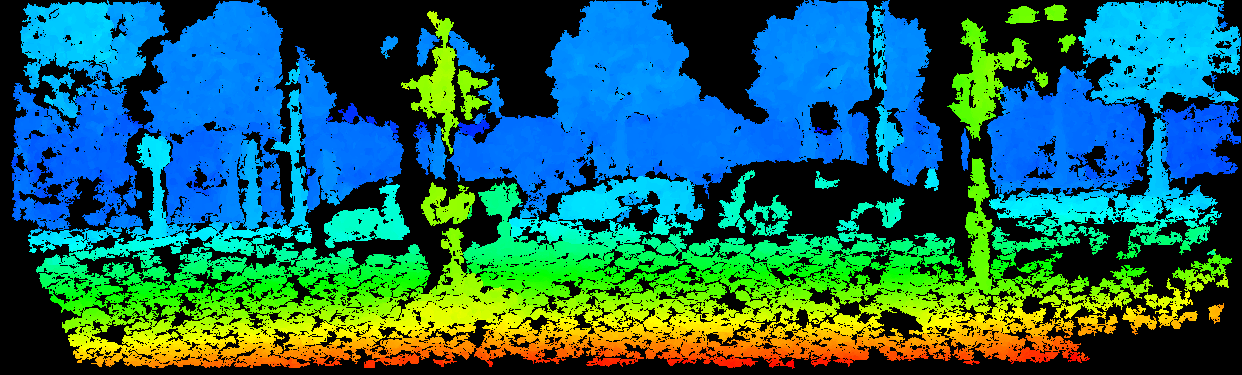}
		\end{subfigure}
		\begin{subfigure}[c]{0.475\columnwidth}
			\includegraphics[width=1\textwidth]{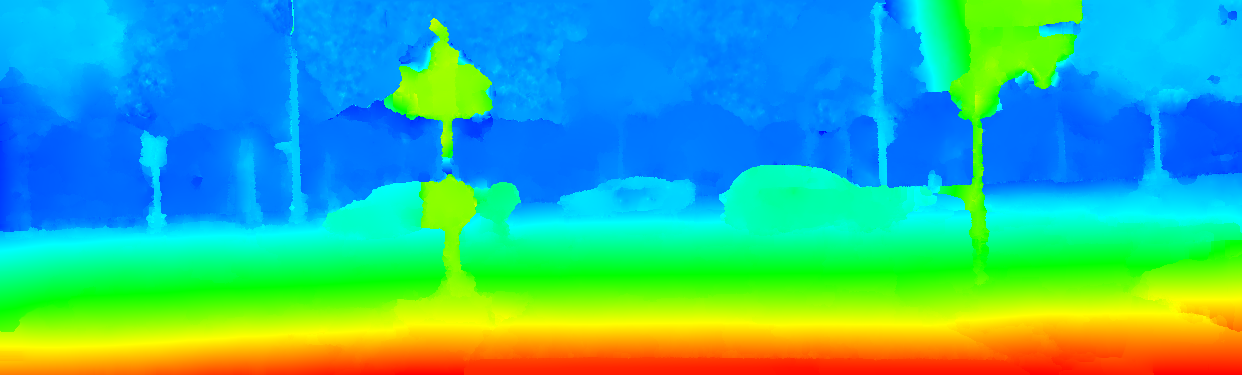}%
			\raisebox{2pt}{\makebox[0pt][r]{\bf c)~}}%
		\end{subfigure}
	\end{center}
	\caption{Sparse correspondences (left) and dense interpolation (right). Optical flow {\bf (a)} and disparities at $t_0$ {\bf (b)} and $t_1$ {\bf (c)}.}
	\label{fig:interpolation}
\end{figure}

\paragraph*{Consistency Check.}
The matching procedure yields a dense map of scene flow correspondences across all images. However, many of the correspondences are wrong because of occlusions, out-of-bounds motion or simply because of mismatching due to challenging image conditions. To remove these outliers, we perform a two-step consistency check. First, we compute an inverse scene flow field for which the reference image is the right image at time $t_1$. Temporal order as well as points of view are swapped. Everything else remains as explained above. During consistency check, optical flow and both disparity maps for each pixel are compared to the corresponding values of the inverse scene flow field. If either difference exceeds a consistency threshold $\tau_c$ in image space, the scene flow vector gets removed. Secondly, we form small regions of the remaining pixels as in \cite{bailer2015flow}, where a pixel is added to a region if it has approximately the same scene flow vector. Afterwards, we check if we could add one of the already removed outliers in the neighborhood following the same rule. If this is possible and the region is smaller than $s_c$ pixels, we remove the whole region. This way we obtain the filtered final scene flow correspondences of high accuracy and very few outliers (cf. \cref{tab:variants}). 
Because most of the times the joint filtering of the matches removes more disparity values than necessary, we fill up gaps with additional values. These values are the result of a separate consistency check for the disparity matches only. For the separate check we compute a second disparity map with \gls*{sgm} \cite{hirschmuller2008SGM} and use the same threshold $\tau_c$ as before. The additional disparity values that are retrieved this way are as accurate as the one from our standard consistency check but much denser which is shown in \cref{fig:interpolation,tab:variants}.

\subsection{Dense Interpolation} \label{sec:interpolation}
\paragraph*{Sparsification.}
Before interpolating the filtered scene flow field to recover full density, an additional sparsification step is performed. This helps to extend the spatial support of the neighborhood during interpolation and speeds up the whole process \cite{bailer2015flow}. For non-overlapping $3 \times 3$ blocks, we select the match with the lowest consistency error during filtering only. The remaining matches are called seeds with respect to the interpolation.

\paragraph*{Interpolation Boundaries.}
A crucial part of the interpolation is the estimation of scene flow boundaries. While \cite{bailer2015flow,revaud2015epic} approximate motion boundaries for optical flow with a texture-agnostic edge detector \cite{dollar2013sed}, our edge detector is trained on semantic boundaries. We find that this models geometric boundaries as well as motion boundaries much better than image edges and is much more robust to lighting, shadows, and coarse textures. To do so, we have gathered about 400 images of the KITTI data set from \cite{rwthsemantics,ros2015offline,xu2016multimodal} that have been labeled with semantic class information. Within these images we have merged semantic classes that in general neither align with geometric nor motion discontinuities, \eg lane markings and road, or pole and panel. The boundaries between the remaining semantic labels are used as binary edge maps to train our edge detector. To this end, we utilize the framework of \gls*{sed} \cite{dollar2013sed} and train a random forest with the same parameters as in their paper, except for the number of training patches. We sample twice as many positive and negative patches during training because we use a bigger data set with images of higher resolution. The impact of the novel boundary detector will be evaluated in \cref{sec:eval}.

\begin{figure}[t]
	\begin{center}
		\begin{subfigure}[c]{0.75\columnwidth}
			\includegraphics[width=1\textwidth]{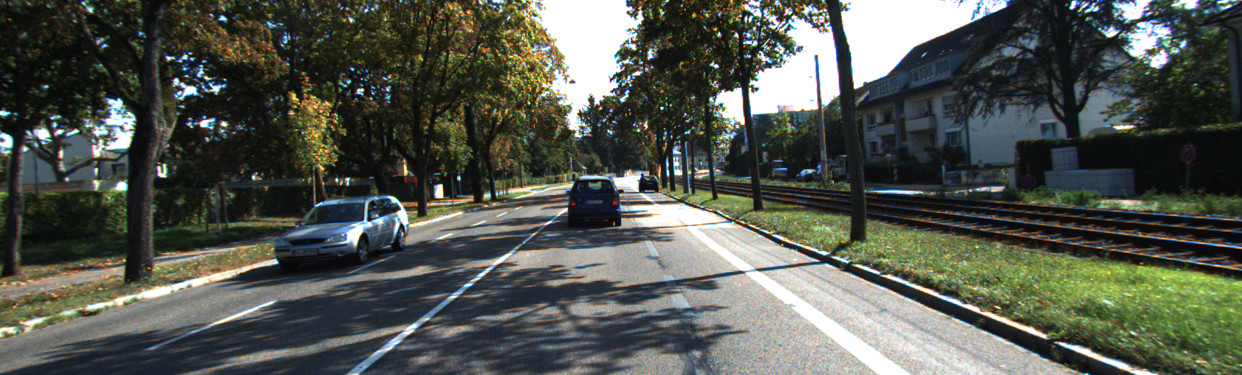}%
			\raisebox{2pt}{\makebox[0pt][r]{\textcolor{white}{\bf a)~}}}%
			\vspace{1mm}%
		\end{subfigure}
		\begin{subfigure}[c]{0.75\columnwidth}
			\includegraphics[width=1\textwidth]{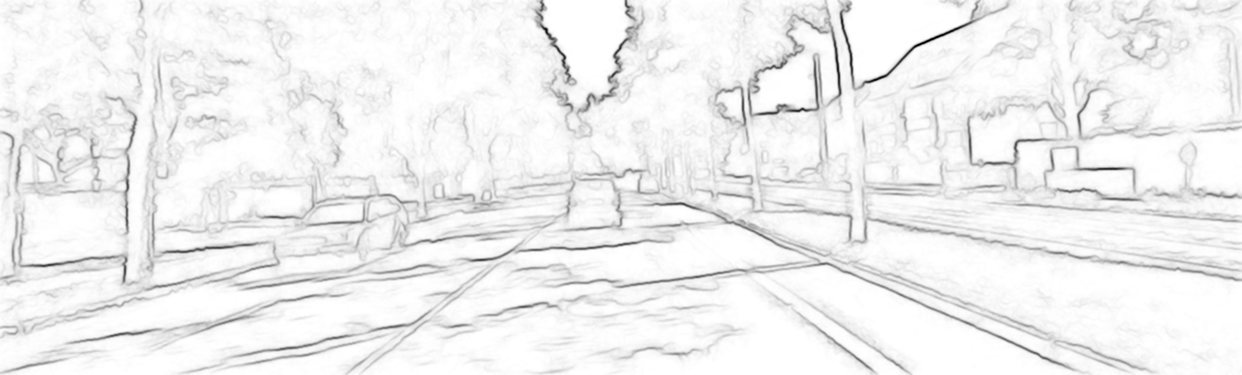}%
			\raisebox{2pt}{\makebox[0pt][r]{\bf b)~}}%
			\vspace{1mm}%
		\end{subfigure}
		\begin{subfigure}[c]{0.75\columnwidth}
			\includegraphics[width=1\textwidth]{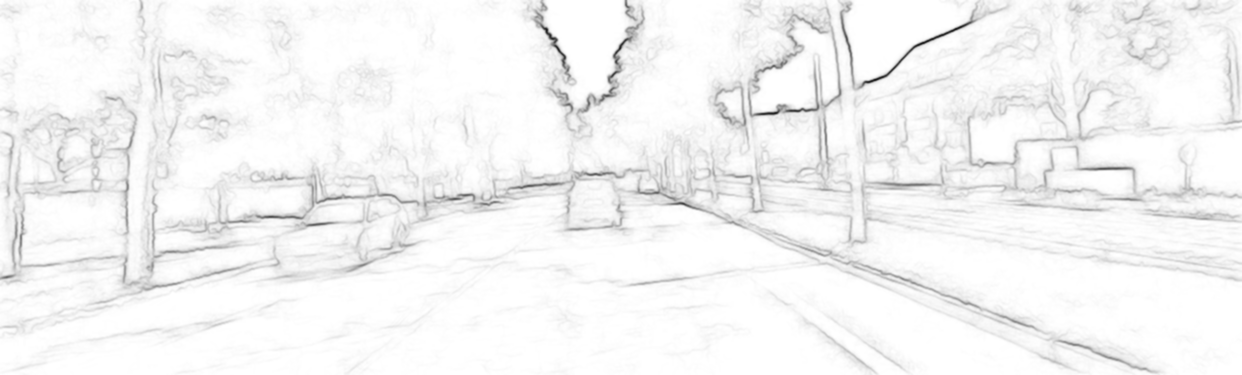}%
			\raisebox{2pt}{\makebox[0pt][r]{\bf c)~}}%
		\end{subfigure}
	\end{center}
\caption{Whereas \gls*{sed} \cite{dollar2013sed} {\bf (b)} detects all image boundaries, our new boundary detector {\bf (c)} suppresses lane markings and shadows.}
\label{fig:edges}
\end{figure}

\paragraph*{Interpolation Models.}
For the interpolation of geometry and motion, we use two different models. Both parts are interpolated separately which leads to a more accurate reconstruction of the scene. This is due to the fact that the separate consistency check for disparity leaves more geometric matches where motion would leave image boundaries. Suppose a local, boundary-aware neighborhood of seeds is given for each unknown scene flow vector $\mathbf{\hat{u}}$ at pixel $\hat{p}$ for geometric and motion seeds respectively, $\mathcal{N}_{geo}$ and $\mathcal{N}_{motion}$. The depth of pixel $\hat{p}$ is reconstructed by fitting a plane $E(\hat{p}): a_1 x + a_2 y + a_3 = d_0$ through all seeds of the neighborhood $\mathcal{N}_{geo}$. This is done by solving a linear system of equations for all neighboring seed points $p_g$ for which the disparity values are known, using weighted least squares. The weights for each seed are obtained from a Gaussian kernel $g(D)= \exp{(-\alpha D)}$ on the distance $D(\hat{p},p_g)$ between target pixel and seed. The missing disparity value of $\hat{p}$ is obtained by plugging the coordinates of $\hat{p}$ into the estimated plane equation. In a similar fashion, but using a neighborhood of motion seeds $\mathcal{N}_{motion}$, the missing 3D motion is obtained by fitting an affine 3D transformation $\mathbf{x_1} = \mathbf{A} \mathbf{x_0} + \mathbf{t}$ using weighted least squares on all motion seeds $p_m$. Where $\mathbf{x_t} = (x_t,y_t,z_t)^T$ are the 3D world coordinates of motion seed $p_m$ at time $t_0$ and $t_1$, and $\left[\mathbf{A}|\mathbf{t}\right] \in \mathbb{R}^{3 \times 4}$ is the affine 3D transformation of twelve unknowns. The weights are computed by the same Gaussian kernel as for geometric interpolation, but using the distances $D(\hat{p},p_m)$ between the target pixel and the motion seeds. To summarize, for the full reconstruction of scene flow $\mathbf{\hat{u}}=(u,v,d_0,d_1)^T$ at pixel $\hat{p}$, we compute $d_0$ using the plane model $E(\hat{p})$, reproject the point into 3D world space, transform it according to its associated affine transformation $\left[\mathbf{A}|\mathbf{t}\right]$, and project it back to image space to obtain $u$, $v$ and $d_1$.

\paragraph*{Edge-Aware Neighborhood.}
To find the local neighborhoods, we follow the idea of \cite{revaud2015epic} using both their approximations. That is first, the $n$ closest seeds to a pixel $\hat{p}$ are the $n-1$ closest seeds to the closest seed of $\hat{p}$, thus all pixels with the same closest seed share the same local neighborhood. And secondly, the distance between $\hat{p}$ and its closest seed is a constant offset for all neighboring seeds which can be neglected. It is therefore sufficient to find a labeling that assigns each pixel to its closest seed and to find the local neighborhood for each seed. We use the graph-based method of \cite{revaud2015epic} for this. Where the distances between seeds are geodesic distances that are directly based on the edge maps from our boundary detector.

\subsection{Variational Optimization} \label{sec:variational}
To further refine the 3D motion after interpolation, we use variational energy minimization to optimize the objective
\begin{equation} \label{eq:energy}
E(u,v,d') = E_{data}^{flow} + E_{data}^{cross} + \varphi \cdot E_{smooth}
\end{equation}
Motion is represented in image space by optical flow and the change in disparity $d'$. The energy consists of three parts. Two data terms, one temporal correspondence and one cross correspondence, and an adaptively weighted smoothness term for regularization. The data terms use the gradient constancy assumption. Our experiments have shown, that a term for the color constancy assumption can be neglected.
\begin{multline} \label{eq:dataterm}
E_{data}^{\ast}(I_1,I_2,\mathbf{x},\mathbf{w}) = \\
\int_{\Omega} \beta\left(\mathbf{x},\mathbf{w}\right) \cdot \Psi\left( \gamma\cdot\left|\nabla I_2(\mathbf{x}+\mathbf{w})-\nabla I_1(\mathbf{x})\right|^2\right)dx
\end{multline}
\begin{equation} \label{eq:flowdata}
E_{data}^{flow} = E_{data}^{\ast}\left(I_l^0,I_l^1,\mathbf{x},(u,v)^T\right)
\end{equation}
\begin{equation} \label{eq:crossdata}
E_{data}^{cross} = E_{data}^{\ast}\left(I_l^0,I_r^1,\mathbf{x},(u-d_0-d',v)^T\right)
\end{equation}
The data terms do not contribute to the energy if the function
\begin{equation}
\beta\left(\mathbf{x},\mathbf{w}\right) = \begin{cases}
1, &\text{if} \left(\mathbf{x}+\mathbf{w}\right)^T \in \Omega \\
0, &\text{otherwise}
\end{cases}
\end{equation}
indicates that the scene flow is leaving the image domain. The smoothness term 
\begin{equation} \label{eq:smoothness}
E_{smooth} = \int_{\Omega} \Psi\left(\left|\nabla u\right|^2 + \left|\nabla v\right|^2 + \lambda \cdot \left|\nabla d'\right|^2\right)dx
\end{equation}
penalizes changes in the motion field and is weighted by 
\begin{equation} \label{eq:smoothnessweight}
\varphi(\mathbf{x}) = e^{-\kappa B(\mathbf{x})}
\end{equation}
where $B(\mathbf{x})$ is the edge value of our boundary detector at pixel $\mathbf{x}$. All parts use the Charbonnier penalty $\Psi\left(x^2\right) = \sqrt{x^2+\varepsilon^2}$ to achieve robustness. Since the smoothness term rather enforces constancy if $\beta$ for both data terms is zero, we do not optimize the scene flow at pixels where the interpolated scene flow field leaves $\Omega$. Our energy formulation is inspired by \cite{brox2004high,huguet2007variational,wedel2008efficient}.
We use linear approximations of the Euler-Lagrange equations for the objective and apply the framework of Brox \etal \cite{brox2004high} without the coarse-to-fine steps to find a solution by \gls*{sor}.

\begin{figure}[t]
	\begin{center}
		\begin{subfigure}[c]{0.75\columnwidth}
			\includegraphics[width=1\textwidth]{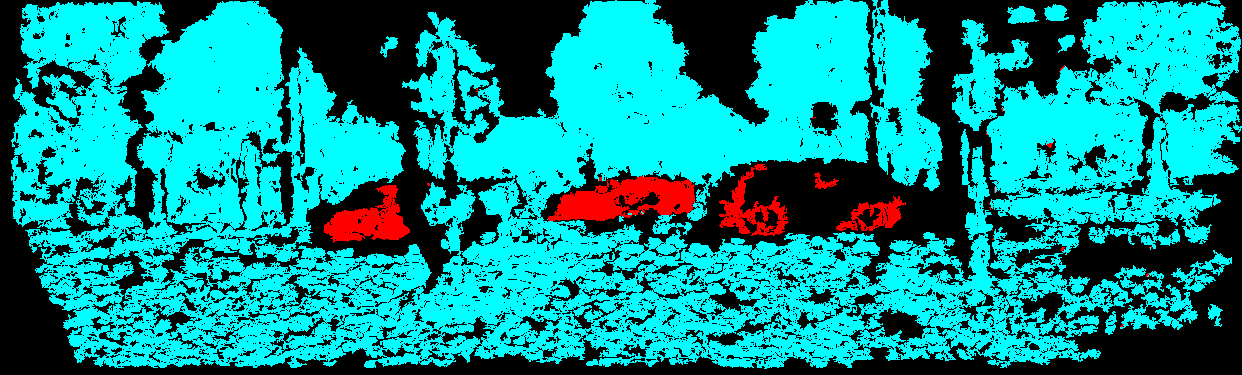}%
			\raisebox{2pt}{\makebox[0pt][r]{\textcolor{white}{\bf a)~}}}%
			\vspace{0.5mm}%
		\end{subfigure}
		\begin{subfigure}[c]{0.75\columnwidth}
			\includegraphics[width=1\textwidth]{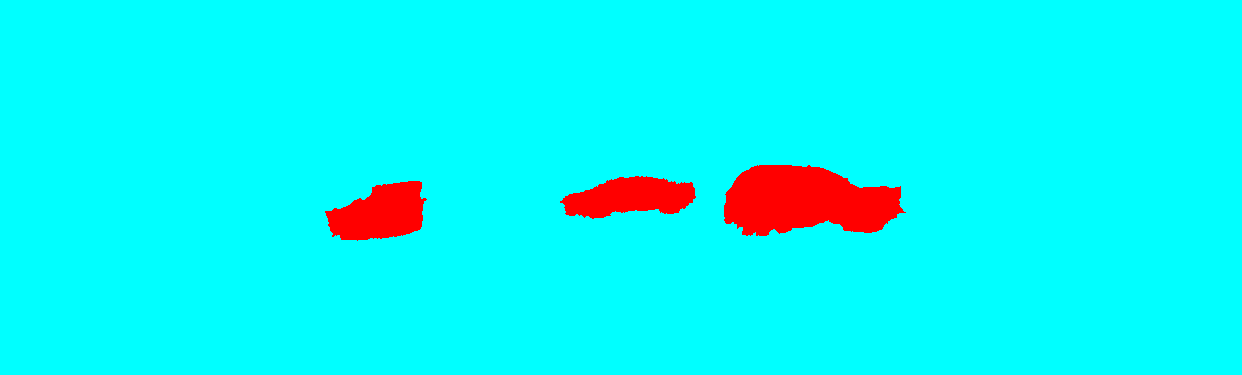}%
			\raisebox{2pt}{\makebox[0pt][r]{\bf b)~}}%
			\vspace{0.5mm}%
		\end{subfigure}
		\begin{subfigure}[c]{0.75\columnwidth}
			\includegraphics[width=1\textwidth]{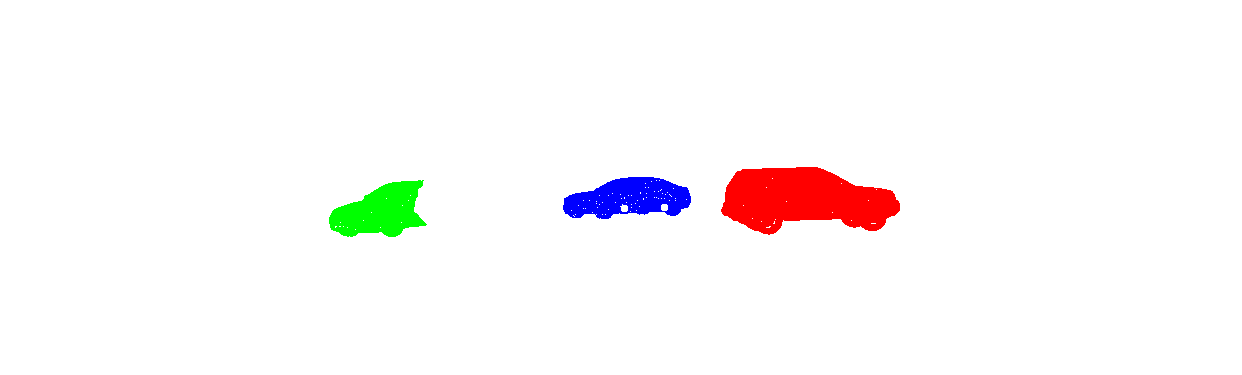}%
			\raisebox{2pt}{\makebox[0pt][r]{\bf c)~}}%
		\end{subfigure}
	\end{center}
	\caption{Example of our motion segmentation. Sparse motion indicators as obtained during ego-motion computation {\bf (a)}, dense segmentation by interpolation {\bf (b)} and moving ground truth objects as provided by KITTI \cite{menze2015object} {\bf (c)}.}
	\label{fig:segmentation}
\end{figure}

\section{Ego-Motion Model} \label{sec:egomotion}
In \cref{sec:eval} we will show that our approach as described so far achieves results comparable to state-of-the-art. For the special challenges of the KITTI data set, we make an additional, optional assumption to further improve the performance of SceneFlowFields. Following \cite{taniai2017fsf}, we argue that most parts of a scene are static and thus that the 3D motion for these areas is fully determined by the ego-motion of the observer. Given the ego-motion and a motion segmentation into static and dynamic areas, we apply the inverse ego-motion to all static points in the scene. 
Using our matching and interpolation scheme, both can be easily estimated with almost no additional effort.

\paragraph*{Ego-Motion Estimation.}
The filtered scene flow field before interpolation provides very accurate matches across all images. We compute 3D-2D correspondences between the reference frame and the temporally subsequent frame by triangulation with the stereo matches. We limit the depth of these correspondences to $35$ meters because disparity resolution for farther distances gets too inaccurate. This way, we obtain a Perspective-n-Point problem, which we solve iteratively using Levenberg-Marquardt and RANSAC to find the relative pose between the left cameras at time $t_0$ and $t_1$ by minimizing the re-projection error of all correspondences. For RANSAC, we consider a correspondence an outlier if the re-projection error is above $1$ pixel. After first estimation, we recompute the set of inliers with a relaxed threshold of $3$ pixels and re-estimate the pose $P=\left[R|t\right] \in \mathbb{R}^{3 \times 4}$. The two stage process helps to avoid local optima and to find a trade-off between diverse and robust correspondences.

\paragraph*{Motion Segmentation.}
An initial sparse motion segmentation can directly be obtained as side product of the ego-motion estimation. Outliers in the correspondences are considered in motion, while points in conformity with the estimated ego-motion are marked as static. We use our boundary-aware interpolation to compute a dense segmentation (cf. \cref{fig:segmentation}). Pixels labeled as moving are spread up to the object boundaries within they are detected. Because the segmentation is only a binary labeling, no complex interpolation model is needed. An unknown pixel gets assigned with the weighted mean of its local neighborhood. The weights are again based on the geodesic distances between matches. This interpolation method is similar to the Nadaraya-Watson estimator in \cite{revaud2015epic}. The interpolated motion field is then thresholded to obtain a dense, binary motion segmentation. The quality of this segmentation is evaluated in \cref{sec:eval}. 
Finally, the inverse estimated ego-motion is applied to all points that are labeled as static.

\section{Experiments and Results} \label{sec:eval}

\begin{table*}[t]
	\begin{center}
		\resizebox{\textwidth}{!}{\begin{tabular}{c | c | c | c | c | c | c | c | c | c | c | c | c | c | c | c}
{\bf Rank} & {\bf Method} & {\bf dual} & {\bf D1-bg} & {\bf D1-fg} & {\bf D1-all} & {\bf D2-bg} & {\bf D2-fg} & {\bf D2-all} & {\bf Fl-bg} & {\bf Fl-fg} & {\bf Fl-all} & {\bf SF-bg} & {\bf SF-fg} & {\bf SF-all} & {\bf Runtime} \Bstrut\\ 
\hline
1 & \acrshort*{prsm} \cite{vogel2015PRSM} &   & {\bf 3.02} & 10.52 & {\bf 4.27} & {\bf 5.13} & {\bf 15.11} & {\bf 6.79} & {\bf 5.33} & 13.40 & {\bf 6.68} & {\bf 6.61} & 20.79 & {\bf 8.97}  & 300 s \Tstrut\\
2 & OSF+TC \cite{neoral2017object} &   & 4.11  & {\bf 9.64} & 5.03  & 5.18 & 15.12 & 6.84 & 5.76 & {\bf 13.31} & 7.02 & 7.08 & {\bf 20.03} & 9.23 & 3000 s\\
3 & OSF \cite{menze2015object} & x & 4.54  & 12.03  & 5.79 & 5.45 & 19.41 & 7.77  & 5.62  & 18.92  & 7.83  & 7.01  & 26.34  & 10.23  & 3000 s\\
4 & FSF+MS \cite{taniai2017fsf} &   & 5.72  & 11.84  & 6.74  & 7.57  & 21.28  & 9.85  & 8.48  & 25.43  & 11.30  & 11.17  & 33.91  & 14.96  & {\bf 2.7 s} \\
5 & CSF \cite{lv2016CSF} & x & 4.57  & 13.04  & 5.98  & 7.92  & 20.76  & 10.06  & 10.40  & 25.78  & 12.96  & 12.21  & 33.21  & 15.71  & 80 s \\
6 & {\bf SceneFlowFields} & x & 5.12 & 13.83 & 6.57 & 8.47 & 21.83 & 10.69 & 10.58 & 24.41 & 12.88 & 12.48 & 32.28 & 15.78 & 65 s \\
7 & PRSF \cite{vogel2013PRSF} & x & 4.74  & 13.74  & 6.24  & 11.14  & 20.47  & 12.69  & 11.73  & 24.33  & 13.83  & 13.49  & 31.22  & 16.44  & 150 s \\
8 & \acrshort*{sgm}+SF \cite{hirschmuller2008SGM,hornacek2014sphereflow} & x & 5.15 & 15.29 & 6.84 & 14.10 & 23.13 & 15.60 & 20.91 & 25.50 & 21.67 & 23.09 & 34.46 & 24.98 & 2700 s \\
9 & PCOF-LDOF \cite{derome2016prediction} & x & 6.31 & 19.24 & 8.46 & 19.09 & 30.54 & 20.99 & 14.34 & 38.32 & 18.33 & 25.26 & 49.39 & 29.27 & 50 s \\
10 & PCOF+ACTF \cite{derome2016prediction} & x & 6.31 & 19.24 & 8.46 & 19.15 & 36.27 & 22.00 & 14.89 & 60.15 & 22.43 & 25.77 & 67.75 & 32.76 & (0.08 s) \\
		\end{tabular}}
	\end{center}
	\caption{Results on the KITTI Scene Flow Benchmark \cite{menze2015object}. The column \textit{dual} indicates whether only two frame pairs are used by this method. Run times in parentheses are using a GPU. We achieve the third best result among all dual-frame methods. Our SceneFlowFields yields especially good results at foreground regions (\textit{SF-fg}).}
	\label{tab:kitti}
\end{table*}
\begin{table*}[t]
	\begin{center}
		\resizebox{0.9\textwidth}{!}{\begin{tabular}{c | c | c | c | c | c | c | c | c | c | c | c | c | c | c}
{\bf Variant} & {\bf D1-bg} & {\bf D1-fg} & {\bf D1-all} & {\bf D2-bg} & {\bf D2-fg} & {\bf D2-all} & {\bf Fl-bg} & {\bf Fl-fg} & {\bf Fl-all} & {\bf SF-bg} & {\bf SF-fg} & {\bf SF-all} & {\bf Density} & {\bf Edges} \Bstrut\\ 
\hline
full+ego & \textbf{5.36} & \textbf{10.85} & \textbf{6.20} & \textbf{7.94} & 18.23 & \textbf{9.51} & \textbf{10.36} & 22.85 & \textbf{12.28} & \textbf{12.04} & 28.31 & \textbf{14.53} & 100.00 \% & \multirow{3}{*}{\rotatebox[origin=c]{90}{Ours}} \Tstrut\\
full & \textbf{5.36} & \textbf{10.85} & \textbf{6.20} & 15.91 & \textbf{18.03} & 16.23 & 22.33 & \textbf{21.69} & 22.23 & 24.78 & \textbf{27.37} & 25.18 & 100.00 \% &  \\
no var & \textbf{5.36} & \textbf{10.85} & \textbf{6.20} & 15.77 & 18.81 & 16.24 & 23.75 & 23.72 & 23.75 & 25.60 & 28.99 & 26.12 & 100.00 \% &  \Bstrut\\
\hline
full+ego & 5.48 & 11.99 & 6.47 & 9.07 & 19.98 & 10.74 & 11.63 & 25.47 & 13.75 & 13.39 & 30.92 & 16.07 & 100.00 \% & \multirow{3}{*}{\rotatebox[origin=c]{90}{\gls*{sed} \cite{dollar2013sed}}} \Tstrut\\
full & 5.48 & 11.99 & 6.47 & 16.90 & 20.80 & 17.50 & 23.57 & 25.22 & 23.82 & 26.14 & 30.71 & 26.84 & 100.00 \% &  \\
no var & 5.48 & 11.99 & 6.47 & 16.73 & 21.38 & 17.44 & 25.00 & 27.04 & 25.32 & 26.93 & 32.16 & 27.73 & 100.00 \% &  \Bstrut\\
\hline
\hline
matches & 1.91 & \textbf{3.74} & 2.18 & 2.48 & 4.08 & 2.71 & 2.10 & 2.78 & 2.20 & 3.87 & 6.24 & 4.21 & 38.82 \% & --\Tstrut\\
disparity & \textbf{1.42} & 4.06 & \textbf{1.82} & -- & -- & -- & -- & -- & -- & -- & -- & -- & \textbf{57.81 \%} & --\\
		\end{tabular}}
	\end{center}
	\caption{Evaluation of the different parts of our method on KITTI \cite{menze2015object} training data. Our new edge detector outperforms \gls*{sed} \cite{dollar2013sed}. The ego-motion model helps greatly to improve overall results. The bottom two rows show the amount of outliers for our sparse correspondences before interpolation. The density is computed with respect to available KITTI ground truth.}
	\label{tab:variants}
\end{table*}

We use the explicit values of the previous sections and the following parameters for all our experiments, even across different data sets. 
For $k=3$ subscales and full resolution we run 12 iterations of propagation and random search.
We use a consistency threshold of $\tau_c=1$ and a minimal region size of $s_c=150$ for the region filter.
During interpolation we use geometry and motion neighborhoods of $160$ and $80$ seeds respectively and $\alpha=2.2$ for the Gaussian kernel to weight the geodesic distances.
For the variational energy minimization we set $\kappa=5$, $\gamma=0.77$, $\lambda=10$ and $\varepsilon=0.001$. We run two outer and one inner iteration in our optimization framework with $30$ iterations for the \gls*{sor} solver using a relaxation factor of $\omega=1.9$. We threshold the interpolated motion field with $\tau_S=0.4$ to obtain a binary segmentation when applying the ego-motion model.

\begin{figure*}[t]
	\begin{center}
		\begin{subfigure}[c]{0.3\textwidth}
			\includegraphics[width=1\textwidth]{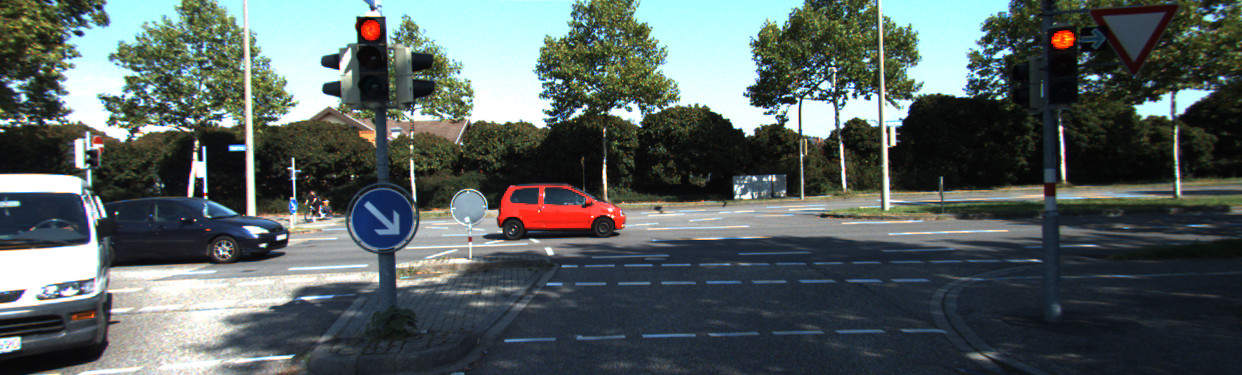}%
			\vspace{1.5mm}%
		\end{subfigure}
		\begin{subfigure}[c]{0.3\textwidth}
			\rule{1\textwidth}{0pt}
		\end{subfigure}
		\begin{subfigure}[c]{0.3\textwidth}
			\includegraphics[width=1\textwidth]{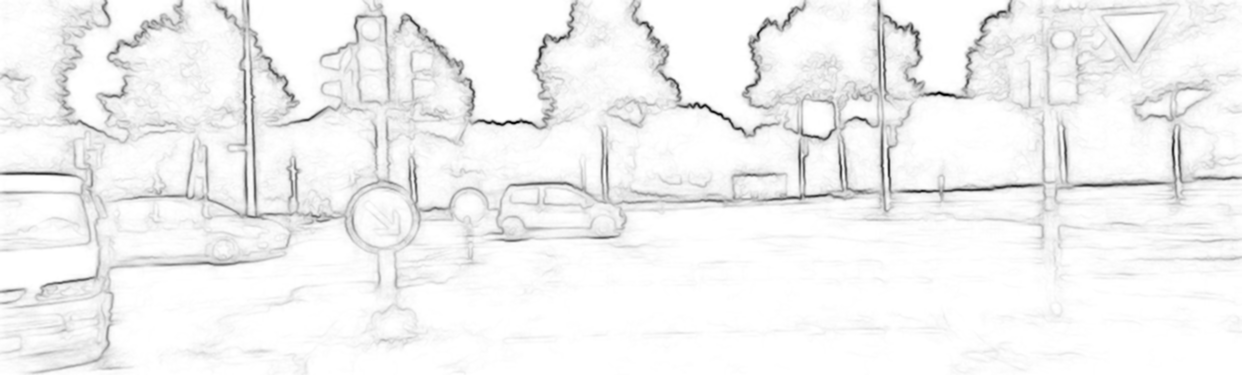}%
			\vspace{1.5mm}%
		\end{subfigure} \\%
		\begin{subfigure}[c]{0.3\textwidth}
			\centering
			\acrshort*{prsm} \cite{vogel2015PRSM} -- multi-frame
		\end{subfigure}
		\begin{subfigure}[c]{0.3\textwidth}
			\centering
			OSF \cite{menze2015object} -- dual-frame
		\end{subfigure}
		\begin{subfigure}[c]{0.3\textwidth}
			\centering
			SceneFlowFields (ours) -- dual-frame
		\end{subfigure} \\%
		\begin{subfigure}[c]{0.3\textwidth}
			\includegraphics[width=1\textwidth]{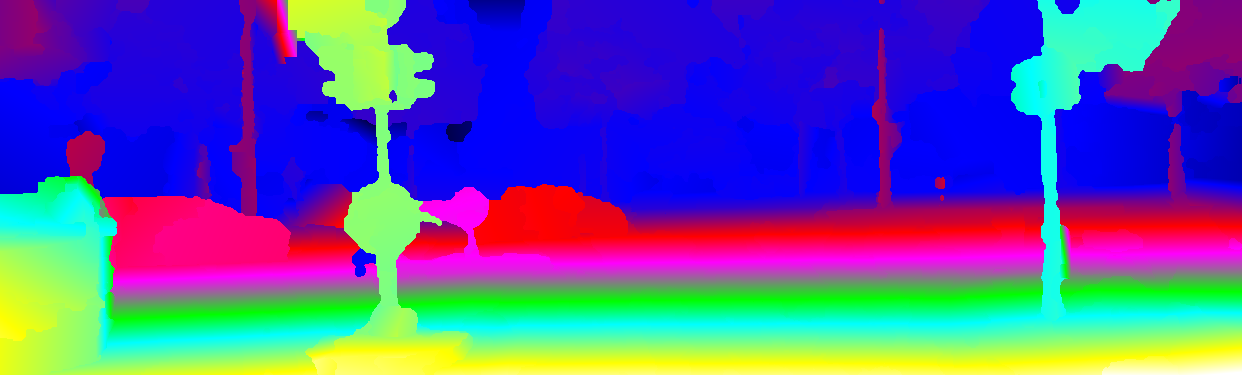}%
			\vspace{0.5mm}%
		\end{subfigure}
		\begin{subfigure}[c]{0.3\textwidth}
			\includegraphics[width=1\textwidth]{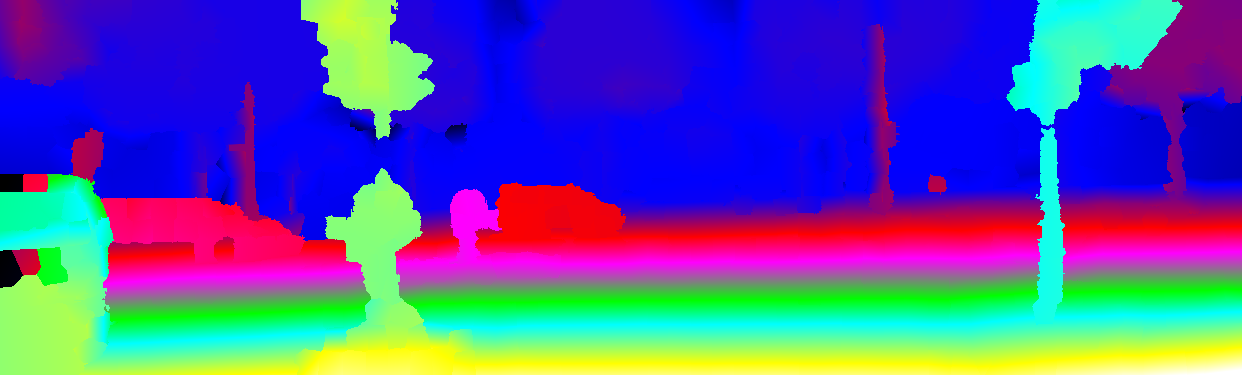}%
			\vspace{0.5mm}%
		\end{subfigure}
		\begin{subfigure}[c]{0.3\textwidth}
			\includegraphics[width=1\textwidth]{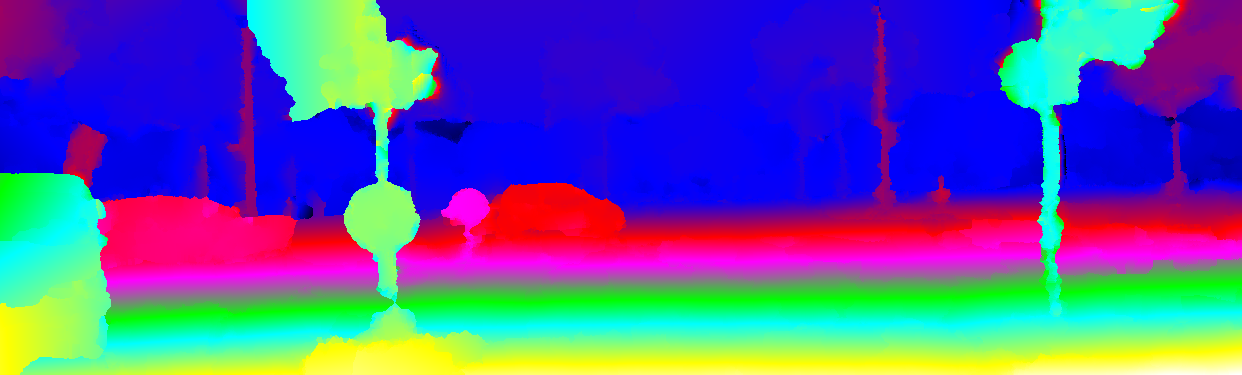}%
			\raisebox{2pt}{\makebox[0pt][r]{\bf a)~}}%
			\vspace{0.5mm}%
		\end{subfigure}
		\begin{subfigure}[c]{0.3\textwidth}
			\includegraphics[width=1\textwidth]{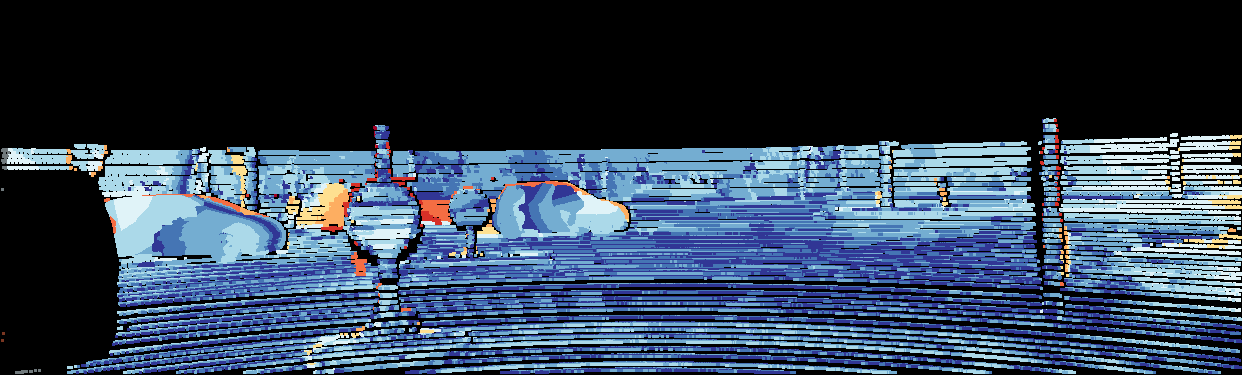}%
			\vspace{0.5mm}%
		\end{subfigure}
		\begin{subfigure}[c]{0.3\textwidth}
			\includegraphics[width=1\textwidth]{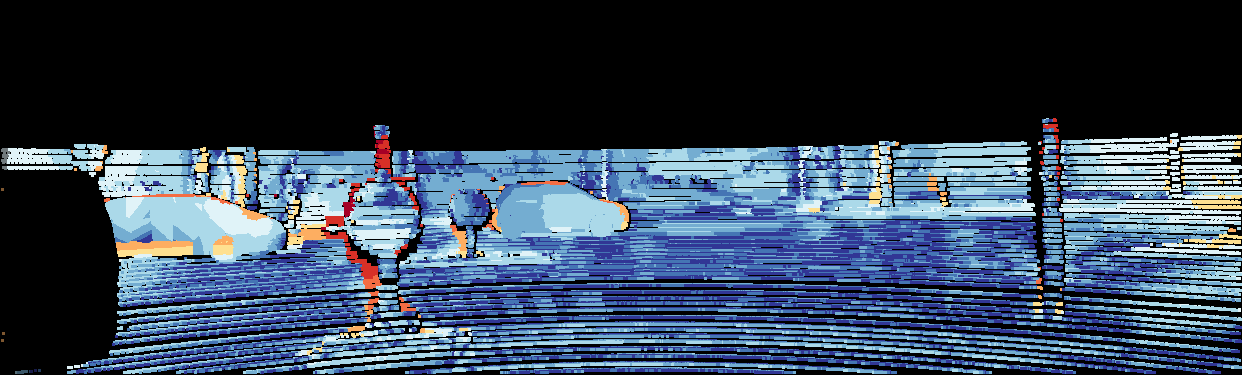}%
			\vspace{0.5mm}%
		\end{subfigure}
		\begin{subfigure}[c]{0.3\textwidth}
			\includegraphics[width=1\textwidth]{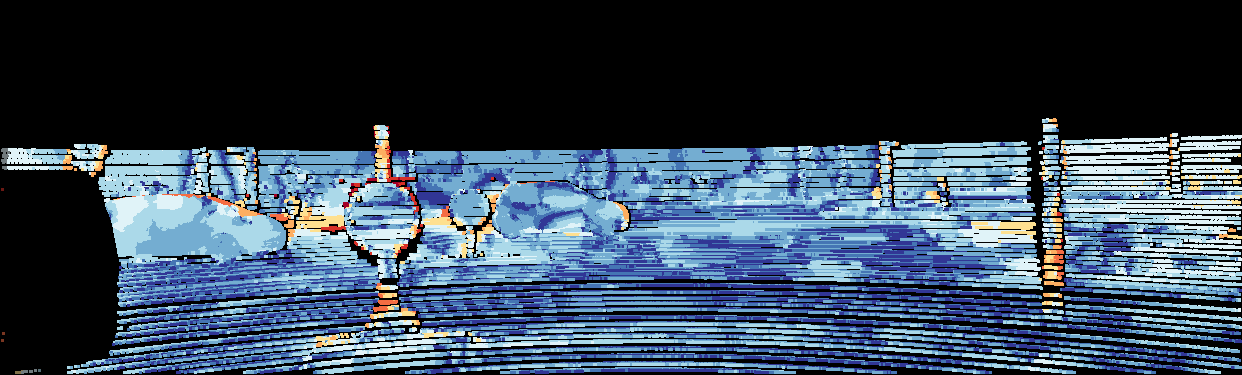}%
			\raisebox{2pt}{\makebox[0pt][r]{\textcolor{white}{\bf b)~}}}%
			\vspace{0.5mm}%
		\end{subfigure}
		\begin{subfigure}[c]{0.3\textwidth}
			\includegraphics[width=1\textwidth]{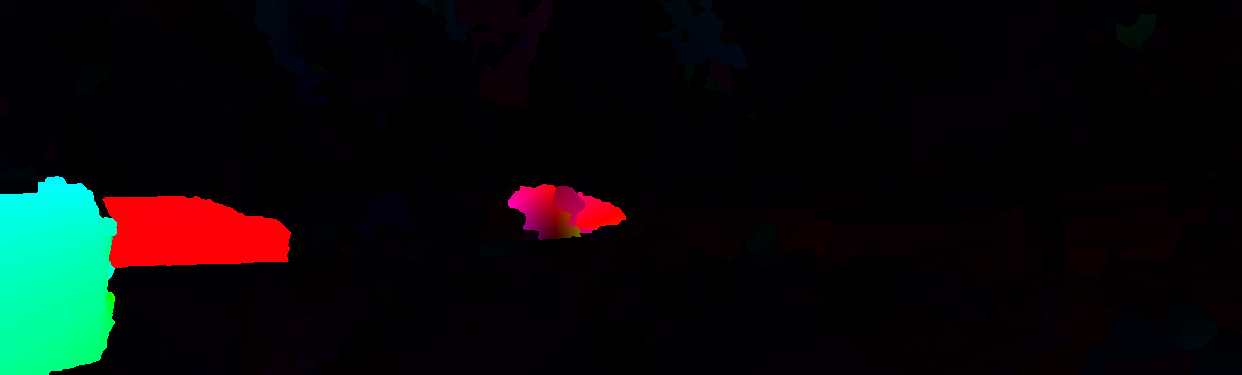}%
			\vspace{0.5mm}%
		\end{subfigure}
		\begin{subfigure}[c]{0.3\textwidth}
			\includegraphics[width=1\textwidth]{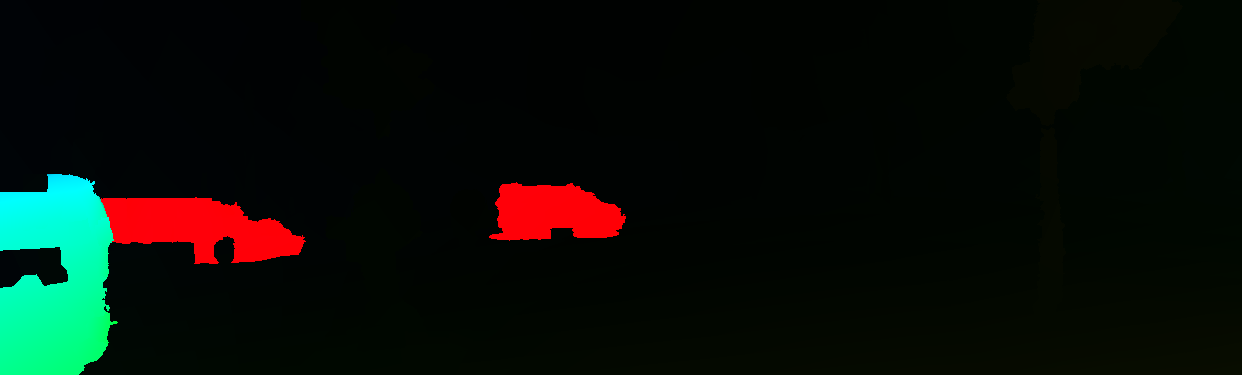}%
			\vspace{0.5mm}%
		\end{subfigure}
		\begin{subfigure}[c]{0.3\textwidth}
			\includegraphics[width=1\textwidth]{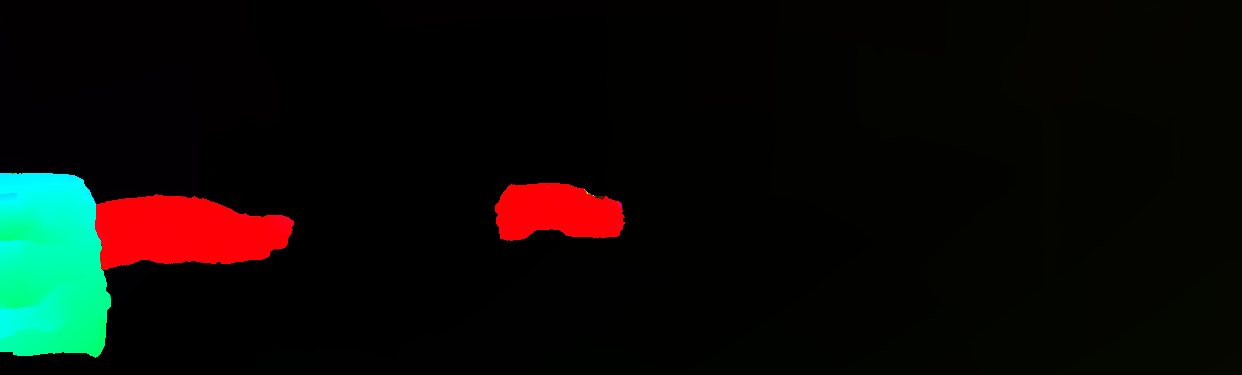}%
			\raisebox{2pt}{\makebox[0pt][r]{\textcolor{white}{\bf c)~}}}%
			\vspace{0.5mm}%
		\end{subfigure}
		\begin{subfigure}[c]{0.3\textwidth}
			\includegraphics[width=1\textwidth]{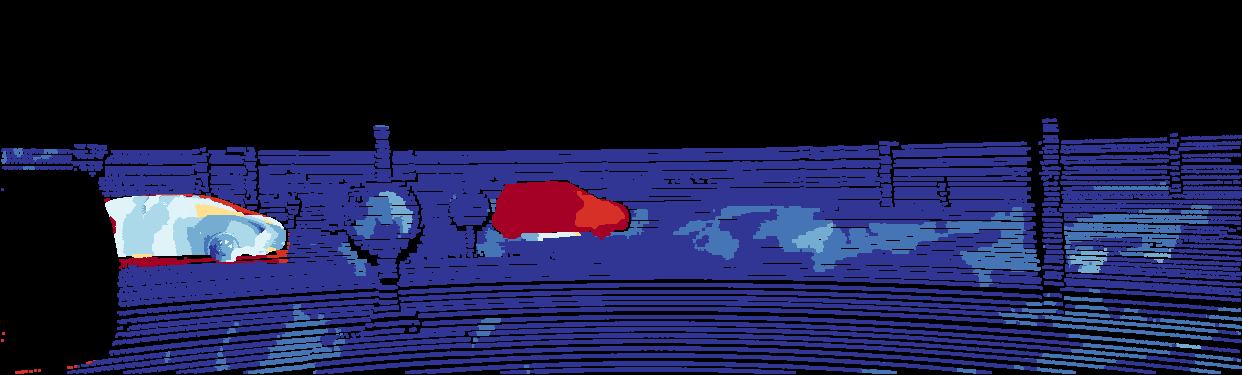}
		\end{subfigure}
		\begin{subfigure}[c]{0.3\textwidth}
			\includegraphics[width=1\textwidth]{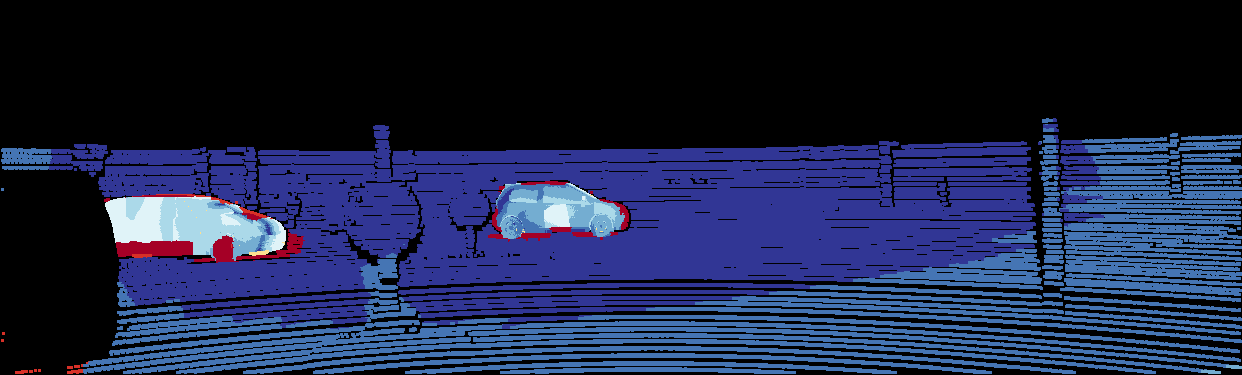}
		\end{subfigure}
		\begin{subfigure}[c]{0.3\textwidth}
			\includegraphics[width=1\textwidth]{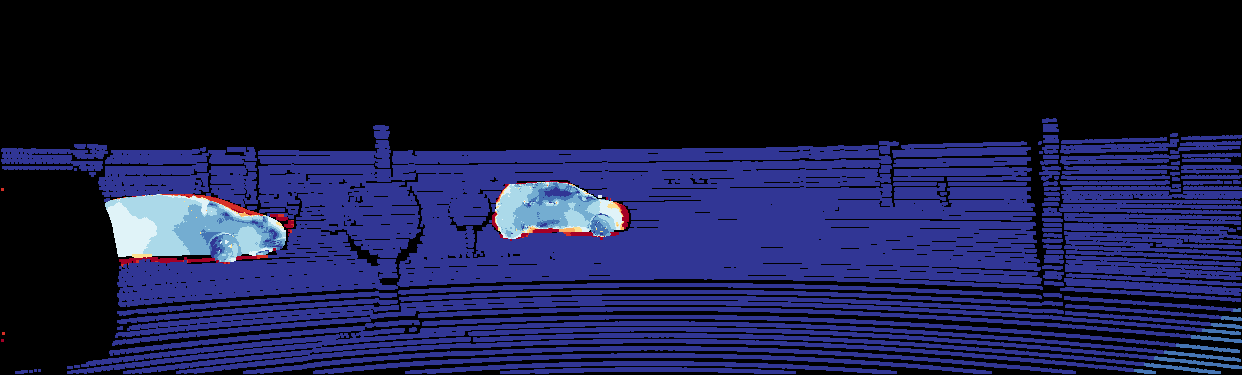}%
			\raisebox{2pt}{\makebox[0pt][r]{\textcolor{white}{\bf d)~}}}%
		\end{subfigure}
	\end{center}
	\caption[Exemplary visual comparison on KITTI Scene Flow Benchmark]{Exemplary visual comparison on KITTI Scene Flow Benchmark \cite{menze2015object}. We show disparity {\bf (a)} and optical flow {\bf (c)} results along with the corresponding error maps {\bf (b)} and {\bf (d)} for \acrshort*{prsm} \cite{vogel2015PRSM}, OSF \cite{menze2015object} and our SceneFlowFields. We accurately detect moving objects and reconstruct sharp boundaries. More examples are visualized in the supplementary video and on the public homepage of KITTI.}
	\label{fig:comparision}
\end{figure*}

\subsection{Boundary Detection} \label{sec:edges}

To test the impact of our motion boundary detector, we evaluate different variants of our method twice. Once, using standard edge detection as in \cite{dollar2013sed}, and a second time using our structured random forest trained on semantic edges. The results are compared in \cref{tab:variants}. The major improvements are visualized in \cref{fig:edges}. High image gradients at lane markings or shadows (especially shadows of vehicles) are effectively suppressed when using our boundary detector, while at the same time it accurately detects all kinds of objects. This helps greatly to smoothly recover the street surface during interpolation, sharpen discontinuities in depth and motion in general, and allows for accurate boundaries when interpolating the motion segmentation.

\subsection{KITTI Scene Flow Benchmark} \label{sec:kitti}
Our main experiments are taken out on the KITTI Scene Flow Benchmark \cite{menze2015object}. The results of our public submission are presented in \cref{tab:kitti} where we compare to \mbox{state-of-the-art} methods. At the time of writing, our method was ranked 6\textsuperscript{th} and achieved the 3\textsuperscript{rd} best result out of all dual-frame methods while at the same time being considerably faster than the top three performing methods (cf. \cref{tab:kitti}). That our method generalizes better to other data sets is shown in \cref{sec:sintel} where we often outperform the best dual-frame method \cite{menze2015object}. In \cref{fig:comparision}, we give an visual example of our results and compare it to the two top performing methods in dual-frame \cite{menze2015object} and multi-frame \cite{vogel2015PRSM} categories. It can be seen that our interpolation produces very sharp edges. This in combination with our matching method helps to obtain accurate scene flow, especially for (moving) objects. Methods with comparable overall performance on KITTI (\cite{lv2016CSF,taniai2017fsf}), perform worse on moving foreground objects than our SceneFlowFields.

Apart from the official evaluation, we test the different components of our method in \cref{tab:variants} to evaluate the effect of each part. We use all training images of the KITTI data set, and evaluate our basic method without the variational optimization (no var), the full basic approach (full) and our method with the optional ego-motion extension (full+ego). Additionally we compute the accuracy and densities with respect to the KITTI ground truth of our sparse scene flow matches (matches) and the separately filtered sparse stereo correspondences (disparity). The variational optimization is primarily useful for optical flow and foreground. All variants using our improved edge detector outperform their according variant using basic image edges.

Finally, we use the provided object maps of KITTI, to test the performance of our motion segmentation (cf. \cref{fig:segmentation}). To this end, we compute the precision and recall for our binary segmentation. Precision is defined as the percentage of estimated pixels that are correctly labeled as in motion. The recall is the relative amount of ground truth pixels that are labeled as moving and covered by our estimation. Over all frames, we achieve a precision of about 28~\% and a recall of about 83~\%. Most of the missed ground truth foreground pixels belong to objects which are far away and moving parallel to the direction of viewing. This way, the re-projection error of the 3D-2D correspondences during ego-motion estimation is below the threshold. Two remarks have to be considered regarding the precision. First, KITTI only annotates cars that are mostly visible, \ie pedestrians, cyclists, other vehicles, or partly occluded cars are not included in the ground truth, but will be marked as moving if they are in motion. Secondly, since areas that are wrongly classified as dynamic will be filled with our basic scene flow estimation which is still of high quality, we did tune in favor of a high recall.

\subsection{MPI Sintel} \label{sec:sintel}
We claim that our proposed method is very versatile and not at all restricted to any setup. Therefore, we have additionally evaluated SceneFlowFields on MPI Sintel \cite{butler2012sintel} without changing any parameters. The only difference to our evaluation on KITTI is, that we do not use our semantic edge detector which was trained on KITTI imagery, but instead \gls*{sed} to obtain edge maps. We test both, our basic approach and the ego-motion extension (+ego) for our method. All training frames over all but two sequences for which a subsequent frame exists are processed. The \textit{final} rendering passes for all images are used. We measure the percentage of outliers according to the KITTI metric for disparity and optical flow. The sequences \textit{cave\_2} and \textit{sleeping\_1} are left out because they have not been evaluated in \cite{taniai2017fsf} -- to which we want to compare -- due to varying camera parameters. The relative amounts of outliers over all evaluated sequences are given in \cref{tab:sintel} and are compared to \cite{menze2015object,taniai2017fsf,vogel2015PRSM} using the results published by \cite{taniai2017fsf}. Our results can keep up with state-of-the-art scene flow methods, although we have not tuned our method on MPI Sintel. For sequences with close-up, non-rigid motion, \eg \textit{ambush\_7} or \textit{bandage\_1}, our depth estimation even beats the multi-frame scene flow method that is ranked first on KITTI.

\begin{table}[h]
	\begin{center}
		\resizebox{\columnwidth}{!}{\begin{tabular}{c|cccc|ccccc}
		  & \multicolumn{4}{c|}{{\bf Disparity}} & \multicolumn{5}{c}{{\bf Optical Flow}} \\
		 {\bf Sequence} & {\bf \acrshort*{prsm}}  & {\bf OSF} & {\bf FSF} & {\bf Ours} & {\bf \acrshort*{prsm}}  & {\bf OSF} & {\bf FSF} & {\bf Ours} & {\bf +ego} \Bstrut\\
		 \hline
		 Average & 15.99 & 19.84 & {\bf 15.35} & 18.15 & {\bf 13.70} & 28.16 & 18.32 & 29.24 & 22.20 \Tstrut\Bstrut\\
		 \hline
		 alley\_1 & 7.43 & {\bf 5.28} & 5.92 & 8.81 & {\bf 1.58} & 7.33 & 2.11 & 5.94 & 3.95 \Tstrut\\
		 alley\_2 & {\bf 0.79} & 1.31 & 2.08 & 1.73 & 1.08 & 1.44 & 1.20 & 2.85 & {\bf 0.87} \\
		 ambush\_2 & 41.77 & 55.13 & {\bf 36.93} & 51.72 & {\bf 51.33} & 87.37 & 72.68 & 90.92 & 83.84 \\
		 ambush\_4 & 24.09 & 24.05 & {\bf 23.30} & 37.78 & {\bf 41.99} & 49.16 & 45.23 & 60.03 & 42.65 \\
		 ambush\_5 & {\bf 17.72} & 19.54 & 18.54 & 25.52 & 25.23 & 44.70 & {\bf 24.82} & 46.92 & 29.86 \\
		 ambush\_6 & 29.41 & {\bf 26.18} & 30.33 & 37.13 & {\bf  41.98} & 54.75 & 44.05 & 57.06 & 47.65 \\
		 ambush\_7 & 35.07 & 71.58 & 23.47 & {\bf 16.34} & {\bf 3.35} & 22.47 & 27.87 & 13.66 & 7.35 \\ 
		 bamboo\_1 & {\bf 7.34} & 9.71 & 9.67 & 14.53 & {\bf 2.41} & 4.04 & 4.11 & 6.11 & 4.15 \\
		 bamboo\_2 & {\bf 17.06} & 18.08 & 19.27 & 19.89 & {\bf 3.58} & 4.86 & 3.65 & 5.84 & 3.97 \\
		 bandage\_1 & 21.22 & 19.37 & 20.93 & {\bf 16.42} & {\bf 3.30} & 18.40 & 4.00 & 3.82 & 4.03 \\
		 bandage\_2 & 22.44 & 23.53 & 22.69 & {\bf 21.77} & {\bf 4.06} & 13.12 & 4.76 & 10.72 & 9.06 \\
		 cave\_4 & {\bf 4.27} & 5.86 & 6.22 & 6.20 & 16.32 & 33.94 & 14.62 & 15.63 & {\bf 12.95} \\
		 market\_2 & {\bf 5.27} & 6.61 & 6.81 & 6.71 & {\bf 4.77} & 10.08 & 5.17 & 7.11 & 6.09 \\
		 market\_5 & 15.38 & 13.67 & {\bf 13.25} & 26.66 & 28.38 & 29.58 & {\bf 26.31} & 40.77 & 28.87 \\
		 market\_6 & {\bf 8.99} & 10.29 & 10.63 & 14.53 & {\bf 10.72} & 16.39 & 13.13 & 28.92 & 16.69 \\
		 mountain\_1 & 0.42 & 0.78 & 0.23 & {\bf 0.15} & {\bf 3.71} & 88.60 & 17.05 & 90.60 & 89.57 \\
		 shaman\_2 & 25.49 & 28.27 & 24.77 & {\bf 21.13} & {\bf 0.46} & 1.67 & 0.56 & 8.85 & 4.31 \\
		 shaman\_3 & 33.92 & 52.22 & {\bf 27.09} & 35.37 & 1.75 & 11.45 & {\bf 1.31} & 15.91 & 8.51 \\
		 sleeping\_2 & {\bf 1.74} & 2.97 & 3.52 & 3.07 & {\bf 0.00} & 0.01 & 0.02 & 0.61 & 0.03 \\
		 temple\_2 & {\bf 4.92} & 5.54 & 5.96 & 6.98 & {\bf 9.51} & 10.52 & 9.66 & 29.58 & 12.57 \\
		 temple\_3 & 11.04 & 16.62 & 10.65 & {\bf 8.61} & {\bf 32.10} & 81.39 & 62.34 & 72.28 & 49.18 \\
		\end{tabular}}
	\end{center}
	\caption{Results on MPI Sintel \cite{butler2012sintel}. Average outliers show that SceneFlowFields can keep up with state-of-the-art.}
	\label{tab:sintel}
\end{table}

\section{Conclusion} \label{sec:conclusion}
Our novel approach to interpolate sparse matches to a dense scene flow achieves state-of-the-art performance on different data sets. At time of submission SceneFlowFields is ranked third on KITTI and achieves state-of-the-art performance on MPI Sintel. We have shown that a stochastic matching approach works for higher dimensional search spaces and that the applied consistency filters produce very robust correspondences. Boundary-aware interpolation has turned out to be a powerful tool to fill the gaps in the scene flow field due to filtering.
To cope with missing correspondences across the images, we have applied an optional ego-motion model that helps to overcome this issue. 
For future work, we want to improve robustness of the ego-motion or extend SceneFlowFields to use multiple frame pairs.

{\small
\bibliographystyle{ieee}
\bibliography{bib}
}

\end{document}